\documentclass{article}

\usepackage{microtype}
\usepackage{graphicx}
\usepackage{subfigure}
\usepackage{booktabs} 

\usepackage{hyperref}

\usepackage[accepted]{whi2020}

\icmltitlerunning{Investigating Bias in Image Classification using Model Explanations}

\begin{document}

\twocolumn[
\icmltitle{Investigating Bias in Image Classification using Model Explanations}



\icmlsetsymbol{equal}{*}

\begin{icmlauthorlist}
\icmlauthor{Schrasing Tong}{mit}
\icmlauthor{Lalana Kagal}{mit}
\end{icmlauthorlist}

\icmlaffiliation{mit}{Department of Electrical Engineering and Computer Science, Massachusetts Institute of Technology, Cambridge, MA, USA}

\icmlcorrespondingauthor{Schrasing Tong}{st9@mit.edu}
\icmlcorrespondingauthor{Lalana Kagal}{lkagal@csail.mit.edu}

\icmlkeywords{Machine Learning, ICML, Computer Vision, Interpretability, Bias, Fairness, Explanations}

\vskip 0.3in
]



\printAffiliationsAndNotice{} 

\begin{abstract}
We evaluated whether model explanations could efficiently detect bias in image classification by highlighting discriminating features, thereby removing the reliance on sensitive attributes for fairness calculations.
To this end, we formulated important characteristics for bias detection and observed how explanations change as the degree of bias in models change.
The paper identifies strengths and best practices for detecting bias using explanations, as well as three main weaknesses: explanations poorly estimate the degree of bias, could potentially introduce additional bias into the analysis, and are sometimes inefficient in terms of human effort involved. 

\end{abstract}

\section{Introduction}
Computer vision has been employed in an increasingly wide range of applications, some of which have serious consequences~\cite{rauscher2013mammogram} in our daily lives.
However, recent research has shown that learned models could produce biased results for minorities or socially vulnerable groups. 
Some prominent examples include Google's algorithm tagging an African American's photo album under "Gorilla"~\cite{google2015racist} and commercial gender classification systems performing worse for females and people with darker skin color~\cite{buolamwini2018gender}.

To this end, researchers have proposed various definitions of fairness~\cite{zafar2017fairness,hardt2016equality} as desirable forms of equality, often relying on sensitive attributes such as race, gender, age, etc.
Nevertheless, the negative effects of bias in the image domain exceed far beyond traditional understandings of sensitive attributes due to the large amount of information present in an image.
For example, a model could give predictions of doctors based on glasses or basketball players based on number 23 jerseys.
Eliminating such biases remains crucial for devising reliable models that generalize well in real world applications.
Alas, current bias detection approaches mostly rely on equality between outputs of different subsets of data and transfer poorly to the image domain for two main reasons:
(i) higher-level concepts are often inferred, for example from a collection of pixels, rather than stated explicitly as a feature and (ii) since causes of bias are not limited to sensitive attributes, brute force attempts to generate finely labeled data tend to be infeasible in terms of efficiency or even replicate existing bias~\cite{kyriakou2019fairness}. 
In this paper, we analyze how examining explanations on models could help us efficiently detect bias by highlighting the discriminating set of features instead of evaluating results for all possible subgroups.
Our work also assesses whether explanations are useful in practice by tackling an important open task and validates if explanation mechanisms faithfully return the appropriate higher-level concepts, in this case the object(s) causing bias.

We formulated important characteristics for bias detection frameworks and analyzed whether two popular explanation mechanisms, namely Grad-CAM~\cite{selvaraju2017grad} and TCAV~\cite{kim2017interpretability}, satisfy them.
We first curated two datasets for image classification, doctors vs nurses and basketball vs volleyball players, that contain fine labels and balanced sample sizes for sensitive attributes to serve as ground truth. 
We then simulated different degrees of bias by altering the training data composition and generated explanations on these models to observe how explanations change as bias vary.
After proposing best practices for human interpretation and processing explanations accordingly, we derived several metrics as operational definitions of fairness in the absence of stated sensitive attributes and compared against current definitions on the degree of bias (unfairness) detected. 
Our results question previous claims~\cite{kim2017interpretability, selvaraju2017grad} that explanations help with bias detection: Although biased models tend to generate explanations that indicate bias, such explanations were observed for relatively unbiased models as well, making it difficult to estimate the degree of bias accurately.
Furthermore, the process of generating or examining explanations sometimes require significant human effort and could also replicate current biases, decreasing the usefulness and trustworthiness of explanations.
By outlining the challenges and intricacies of using explanations to solve the important task of bias detection, this paper promotes the development of novel explanation mechanisms that are useful in practice. 

\section{Related Work}
Fairness definitions, often resembling conditional probabilistic statements on sensitive attributes, establish a form of equality, for example equal accuracy for men and women~\cite{hardt2016equality}, which bias detection seeks to identify if certain (potentially sensitive) attributes violate.
In the image domain, due to the lack of fine labels, previous works~\cite{buolamwini2018gender,kyriakou2019fairness} on bias detection manually audit~\cite{sandvig2014auditing} the model.
This approach relies on human intuition to correctly spot the attribute inducing bias, requires significant effort in curating well-controlled, representative data with ground truth labels, and potentially introduces additional bias when auditing. 
By leveraging explanations, we seek to efficiently recognize bias by focusing on the decision process.
This approach takes advantage of process fairness~\cite{grgic2016process} and moves away from constraints on sensitive attributes. 

Grad-CAM and saliency maps~\cite{simonyan2013deep} highlight regions in the image that impact the decision significantly. 
TCAV takes a very different approach by reporting the percentage of images in which a predetermined concept contributes heavily to predictions.
The authors of both methods provided examples of explanations that indicate bias.
We observed how explanations change as bias changes and compared against a set of important characteristics to evaluate whether explanations actually help detect bias in practice. 
Our work also relates to studies on human interactions with explanations~\cite{lage2019human} and critiques on the faithfulness of explanations~\cite{adebayo2018sanity}, focusing in the context of bias.

\section{Methods}
To evaluate whether explanations help detect bias in practice and facilitate developments in novel explanation mechanisms, we describe the problem setting and assumptions of a real-world scenario and propose important properties for explanation mechanisms. 
Assume that researchers have access to images of the two classes $Y \in \{0, 1\}$ as well as white-box knowledge of a model $f(X) \in \{0, 1\}$, including all its parameters.
There also exists a list of (potentially sensitive) attributes that are suspected to cause bias, with tolerance ranges depending on the attribute, application, and the consequences of error.
Since manual auditing or generating fine labels are inefficient, the researchers seek to identify the cause and degree of bias from explanations for subsequent bias mitigation.
In this setting, an ideal explanation mechanism should encompass the following:
\begin{itemize}
\item Detect bias causes: Detect the causes of bias such as gender, race, image quality, background setting, etc.
\item Detect bias degree: Detect the degree of bias for each attribute, such as ten percent accuracy difference in performance, to compare with acceptable threshold.
\item Efficiency: For saliency maps, processing each explanation require similar human effort to manual labeling when considering only one attribute. TCAV requires significant human effort in concept curation and lacks efficiency if concepts are context dependent and could not be easily reused.
\item Multiple attribute friendliness: Multiple causes of bias could influence the prediction simultaneously and explanations should faithfully reproduce this.
\item Human understandability: For explanations that require human examination, results should be interpretable to a trained expert and not susceptible to inducing human bias.
\end{itemize}
In this paper, we investigate if the two chosen explanation mechanisms satisfy these properties.

\subsection{Dataset Curation}
We evaluate the efficacy of the proposed approach on two image classification datasets.
The doctor-nurse dataset contains gender as the sensitive attribute and the basketball-volleyball dataset contains both gender and jersey color as sensitive attributes; these attributes only represent a small subset of all possible biases and were chosen because they have prominent features.
A standard example of biased predictions for the doctor-nurse dataset would be to treat both gender and clothing as discriminating features, resulting in lower accuracy predictions for the opposite gender.
Curating datasets with fine labels on sensitive attributes greatly improve the soundness of our analysis: deriving ground truth for attributes allows for comparisons against group fairness definitions and 
balancing sample sizes enables simulating various degrees of bias for one attribute while maintaining the same data composition for other attributes.
Since it would be infeasible to account for all spurious correlations and biases, such as posture, background, or presence of glasses, the datasets should only be considered unbiased for the attributes with fine labels, rather than perfectly unbiased. 

\subsection{Model Training and Explanation Generation}
We trained models with different degrees of bias by altering the training data composition and observed how generated explanations change as bias varies.
Biased models tend to incorporate bias-correlated features into its discriminating set, which would be reflected in generated explanations as explanations that indicate bias.
Despite the promising start, explanations that indicate bias do not necessarily imply incorrect predictions or predictions that would change without the presence of bias.
An explanation mechanism that accurately report both the cause and degree of bias should not generate any explanations that indicate bias for unbiased models, with such explanations monotonically increasing as the degree of bias increases. 
Otherwise, it would be hard to estimate the unknown degree of bias in models in practice without first establishing an unbiased baseline.

\subsection{Explanation Evaluation}
In order to detect bias, generated explanations need to be converted into quantitative metrics with the aid of human interpretation.
Since manually examining explanations may introduce additional human bias into the evaluation process, I describe some best practices and limitations for the two very different forms of explanations generated. 
Unlike domains involving specific legal, medical, or financial knowledge, humans have relatively intuitive understandings of biased features in image classification applications, especially given the cause of bias. 

\subsubsection{Grad-CAM}
Grad-CAM explain the model by highlighting regions that contributed heavily to the class decision, which may or may not coincide with higher-level concepts such as the face, hair, or shirts. 
Explanations likely indicate bias when the focused feature only generalizes to limited subsets of the dataset.
For example, if the model focuses on facial features to determine a person's occupation, the prediction may change for members of different gender or race. 
Deciding directly whether explanations show bias suffers from lack of standardization and we propose that domain experts generate a list of features associated with each sensitive attribute in the context of the task. 
Using gender as an example, the list should include not only common features such as clothing or hair style, but also proxy features like footwear or accessories. 
In scenarios when the highlighted area partially overlap with features in the list, one should carefully establish a threshold independent of the instances of the attribute in question. 
Analyzing explanations generated by Grad-CAM require significant human effort, thus, its strength lies in taking into account multiple potentially biased attributes and their associated features at the same time and only indicating which ones contributed to the decision process.

\subsubsection{TCAV}
TCAV explains the model by computing similarities between classes and a set of predetermined high-level concepts. 
The model contains bias when different instances of the same attribute receive very different similarity scores to a given class.
For example, the doctor class shows bias towards men if the similarity score of the men concept greatly exceed that of the women concept.
Although interpreting TCAV explanations for multiple sensitive attributes require little human effort, constructing the concepts involves manually curating a set of images, which lacks efficiency if concepts could not be reused across different tasks.
Furthermore, each concept actually represents a bag of concepts, since a man could also include a white shirt, an office background, Asian ethnicity, and so on.
To prevent additional bias, one must ensure that features in the concepts do not display unwanted correlations with other concepts or classes.

\subsection{Operational Fairness Definitions}
An important first step to tackling bias in the image domain is to establish an acceptable, unbiased state for the sensitive attributes.
In the absence of fine labels for the attributes in question, explanation-based metrics could serve as an operational form of fairness by approximating group fairness definitions.
Taking into account whether explanations show bias improves upon examining only incorrect predictions for two main reasons: (i) since the subgroup population ratio remain unknown, even when the minority group gets predicted incorrectly more often, the majority group could still dominate the incorrect predictions, and (ii) incorrect predictions may always occur and fundamentally differ from incorrect predictions due to biased reasons. 
Deriving fairness metrics from generated explanations depend on the choice of the mechanisms.

\section{Results}
We first describe details and observations on the datasets and trained models.
The two datasets used in our experiments, doctor-nurse and basketball-volleyball, each consists of 8 subgroups of 200 images, 2 for the classes and 4 for combinations of sensitive attributes, downloaded from Google image search queries.
Since the number of subgroups increases exponentially with the sensitive attributes in consideration, curating well-balanced datasets is likely infeasible in practice; the same applies to controlled datasets for manual auditing or TCAV concept representation.
We manually examined the collected data and removed stylized images and images with both instances of the sensitive attribute present. 

For each dataset and sensitive attribute, we fine-tuned five Alexnet~\cite{krizhevsky2012imagenet} models, pre-trained on Imagenet~\cite{russakovsky2015imagenet}, with different degrees of bias.
Table~\ref{tab:1} displays the model performance for subgroups in the doctor-nurse dataset with gender as sensitive attribute, with skin color, balanced for each subgroup to avoid confounding variables.
The ratio A:B represents instances of the sensitive attribute in class 0, in this case male:female in the doctor class, interchanged to B:A for nurses; tables and figures for other datasets and attributes follow the same trend and are included in the supplementary materials. 
As expected, biased models perform better for subgroups that benefit from the bias and vice versa. 
In the first row, when the doctor class consists of male doctors only, the model learns both clothing and gender as discriminating features and assigns more men as doctors while increasing mistakes for male nurses. 
This trend is evident in columns 1 and 4's descending performance and columns 2 and 3's ascending performance.
The Avg column averages performances for subgroups with equal weight to estimate generalization performance.
Since biased models perform slightly better for the favored subgroups at the expense of a large decrease in other subgroups, the 1:1 unbiased model performs best.
However, when the testing data displays the same bias as the training data, in the w bias column, the trend reverses, with biased models performing better by effectively solving only a simpler problem, in this case classifying between male doctors and female nurses rather than doctors and nurses in general.
Bias detection is critical for developing models that generalize well and we analyze if explanations reflect the performance differences in trained models.

\begin{table}[h]
\vskip 0.1in
\begin{center}
\begin{small}
\begin{sc}
\begin{tabular}{ c|cccc|cc } 
 \toprule
 Ratio & $D_M$ & $D_F$ & $N_M$ & $N_F$ & Avg & w bias \\ 
 \midrule
 1:0 & 81.0 & 62.8 & 57.0 & 77.5 & 69.6 & 79.3 \\ 
 3:1 & 77.8 & 72.9 & 76.6 & 79.1 & 76.6 & 77.5 \\ 
 1:1 & 79.4 & 82.2 & 77.3 & 72.1 & 77.8 & 77.8 \\
 1:3 & 71.4 & 77.5 & 84.4 & 72.1 & 76.4 & 78.7 \\
 0:1 & 66.7 & 86.0 & 85.9 & 53.5 & 73.0 & 86.0 \\
 \bottomrule
\end{tabular}
\end{sc}
\end{small}
\end{center}
\vskip -0.1in
\caption{Classification accuracy in percentage for different subgroups in the doctor-nurse dataset with gender as the sensitive attribute. The ratio A:B represents male:female ratio in the doctor class and female:male ratio in the nurse class.}
\label{tab:1}
\end{table}

\subsection{Bias Detection with Grad-CAM}
\begin{figure}[h]
\vskip 0.1in
\begin{center}
\centerline{\includegraphics[width=\columnwidth]{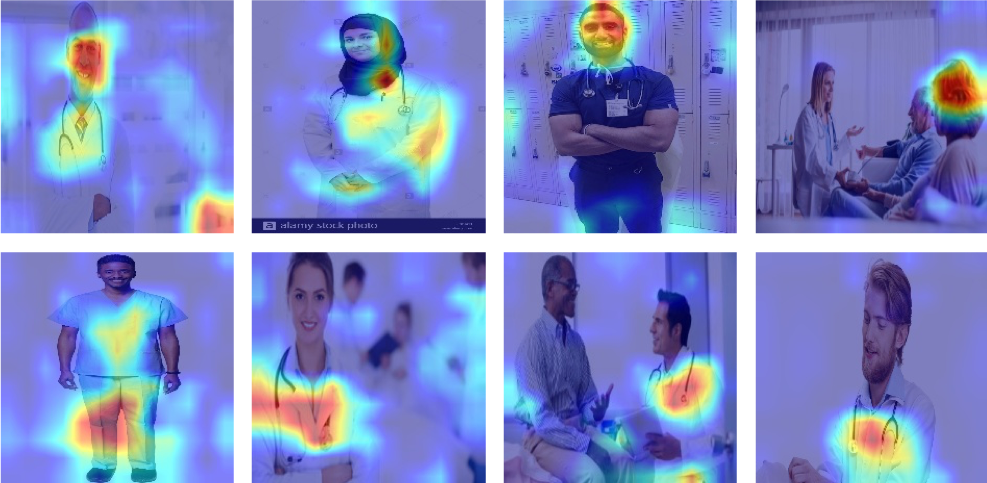}}
\caption{Example explanations for the doctor-nurse dataset with gender as sensitive attribute. Row 1: explanations that show bias (focus face or head) and row 2: unbiased explanations (focus clothing or stethoscope).}
\label{fig1}
\end{center}
\vskip -0.1in
\end{figure}

We first describe the conditions for which an explanation is considered to indicate bias.
For gender bias, we include face, head, hair, and gender specific head-wear as biased features and for jersey color bias, we include jersey and shorts (only if the color matches the jersey).
Figure~\ref{fig1} displays examples of explanations indicating bias and unbiased explanations for the doctor-nurse dataset with gender as sensitive attribute. 
We paid special attention to proxy features: hijabs, first row, second image from left, are associated with women while surgery caps are associated equally with both gender.
When the highlighted region spans multiple features, we considered the explanation unbiased only if none of the regions significantly overlap features on the biased list. 

Using this standard, we manually examined 50 generated explanation for each subgroup and each model and recorded the number of incorrect predictions with explanations showing bias, divided by the total number of explanations showing bias; table~\ref{tab:2} presents our findings for the doctor-nurse dataset with gender as sensitive attribute. 
Although results display promising trends, generated explanations do not accurately estimate the degree of bias or reflect performance disparities.
The number of incorrect predictions based on biased reasons are much higher for subgroups that suffer from the bias.
For example, when all doctors are male, female doctors and male nurses received many incorrect predictions when the explanations focused on facial features.
However, male doctors, the subgroup that benefit from the bias, still sometimes received incorrect predictions due to bias even though they should theoretically all receive correct predictions. 
The unbiased model makes the least number of predictions based on biased reasons.
When the degree of bias increases, the total number of explanations showing bias increases due to the model incorporating more biased features into its discriminating set.
Although the unbiased model shows the least number of such explanations, this number is far from zero, implying that biased features always contribute partially regardless of the degree of bias.
This should be treated as noise and calibrated but doing so in practice is impossible without access to the unbiased baseline model.

\begin{table}[h]
\vskip 0.1in
\begin{center}
\begin{small}
\begin{sc}
\begin{tabular}{ c|cccc|c } 
 \toprule
 Ratio & $D_M$ & $D_F$ & $N_M$ & $N_F$ & Sum \\ 
 \midrule
 1:0 & 3/10 & 7/8 & 8/11 & 1/9 & 19/38 \\ 
 3:1 & 3/9 & 5/9 & 2/11 & 1/11 & 13/40 \\ 
 1:1 & 1/8 & 1/6 & 3/8 & 2/10 & 7/32 \\
 1:3 & 2/8 & 0/6 & 3/9 & 3/13 & 8/36 \\
 0:1 & 2/6 & 1/7 & 2/13 & 9/12 & 14/38 \\
 \bottomrule
\end{tabular}
\end{sc}
\end{small}
\end{center}
\vskip -0.1in
\caption{Number of incorrect predictions with explanations indicating bias/total number of explanations indicating bias for the doctor-nurse dataset with gender as the sensitive attribute.}
\label{tab:2}
\end{table}

\begin{figure}[h]
\vskip 0.1in
\begin{center}
\centerline{\includegraphics[width=\columnwidth]{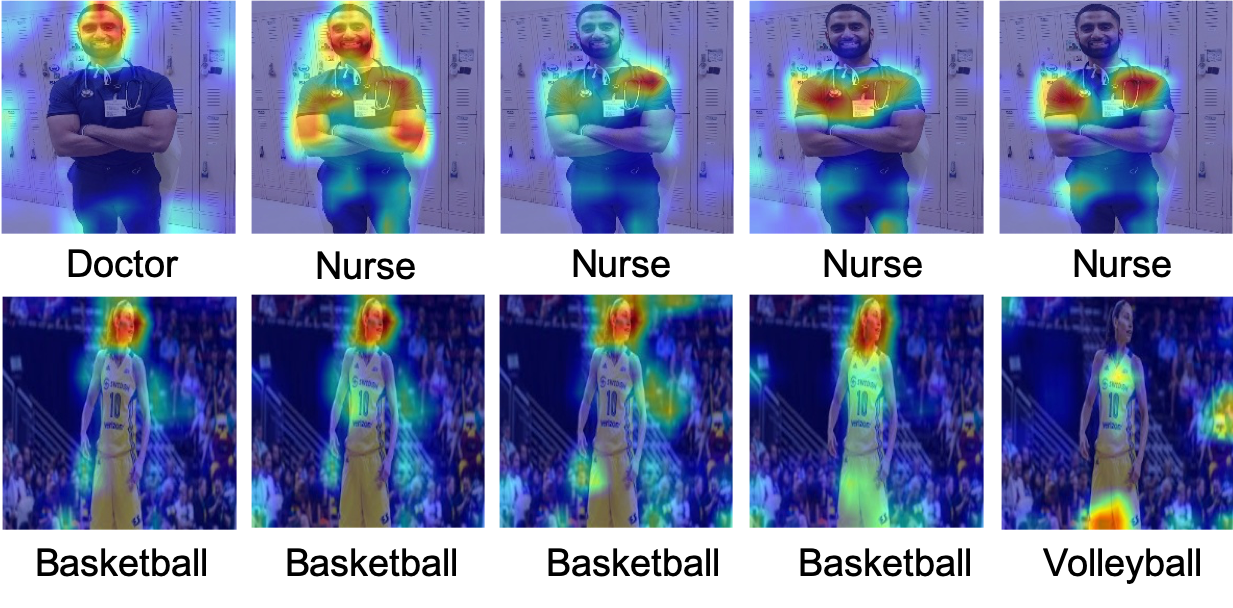}}
\caption{Explanations and predictions for models with different degrees of bias (the leftmost column's model favors male doctors and female volleyball players and the rightmost column male nurses and female basketball players). Ground truth for row 1: male nurse and row 2: female basketball player.}
\label{fig2}
\end{center}
\vskip -0.1in
\end{figure}

Observing how explanations change with the degree of bias provides insights into how explanations sometimes fail, shown in figure~\ref{fig2}.
In the first row, the explanations successfully indicated that the male nurse received incorrect predictions due to bias and correct predictions for justified reasons.
However, although the leftmost and second-leftmost models both favor male doctors and explanations focused significantly on the faces, the images received different predictions.
In row 2, leftmost 2 images, the explanation focused on the female basketball player's face while making the correct prediction despite the model strongly associating females with volleyball players. 

We devised several explanation-based metrics from Table~\ref{tab:2} to detect bias without relying on stated sensitive attributes.
Table~\ref{tab:3} compares the results against the subgroup accuracy disparity, the amount of unfairness in the equal opportunity fairness definition, for the doctor-nurse dataset.
Metric 1, the percentage of incorrect predictions based on biased features, only considers bias that resulted in errors and performs poorly by underestimating the amount of bias.
Metric 2, the total percentage of explanations showing bias, performs better by capturing how bias contributes to predictions in general. 
However, this greatly overestimates bias for the unbiased model when explanations are not perfect.
Metric 3, subgroup differences for the percentage of errors conditional on explanations showing bias implicitly normalizes the contribution of biased features in the unbiased models.
For example, when two subgroups have 5/5 and 5/20, the first subgroup suffers from the bias whereas the second only indicates the unbiased model relies on biased features more. 
Calculating metric 3 requires examining a much higher number of explanations and in general all 3 metrics perform worse than manual labeling and auditing when only one sensitive attribute is concerned.

\begin{table}[h]
\vskip 0.1in
\begin{center}
\begin{small}
\begin{sc}
\begin{tabular}{ c|c|ccc } 
 \toprule 
 Ratio & Unfairness & M1 & M2 & M3 \\ 
 \midrule
 1:0 & 19.4 & 9.5 & 19.0 & 59.6 \\ 
 3:1 & 3.7 & 6.5 & 20.0 & 15.7 \\ 
 1:1 & 2.0 & 3.5 & 16.0 & 10.8 \\
 1:3 & 9.2 & 4.0 & 18.0 & 17.6 \\
 0:1 & 25.9 & 7.0 & 19.0 & 39.3 \\
 \bottomrule
\end{tabular}
\end{sc}
\end{small}
\end{center}
\vskip -0.1in
\caption{Comparison between subgroup accuracy difference (lack of group fairness) and the 3 proposed explanation-based metrics for the doctor-nurse dataset with gender as sensitive attribute.}
\label{tab:3}
\end{table}

The strength of using Grad-CAM to detect bias lies in efficiently detecting multiple attributes at the same time by considering only the attribute shown in the highlighted region.
We trained one model with data compositions equally favoring male players and players wearing red jerseys for the basketball class.
Most of the explanations focused on the jersey and shorts rather than the face or head.
Although the explanations correctly reflected the larger disparity in performance caused by the jersey color bias, estimating the effects of gender was difficult, exacerbated by the sub-optimal performance of explanations in the one attribute scenario.
Since features could be considered biased for one attribute but unbiased for another, the human element in recognizing features and associating them with the appropriate bias in the list plays an important role. 
For instance, most people would consider explanations focusing on the jersey unbiased without specific assumptions but would consider focusing on the face biased.

\subsection{Bias Detection with TCAV}
We represented the gender and jersey color concepts using 100 images each, downloaded from Google image search; the negative examples used to train concepts consist of the opposite gender and players wearing jerseys of all colors respectively. 
In order to detect bias efficiently, concepts should be reused for different tasks, thus, the images we curated contained subjects of different occupations and sports.
However, initial trials with images from the Labeled Faces in the Wild dataset~\cite{huang2008labeled} performed poorly due to differences in cropping (head portrait versus full body), suggesting some restrictions in applicability to other task.
We also noticed that the concepts replicate existing bias by comprising mostly of Caucasians, which is hard to detect without fine labels. 
To study how the model generalizes, different subgroups are represented equally in the target class when computing similarity scores. 
Since we fine-tuned models while freezing the convolutional layer parameters, the same concept activation vectors are reused for models with different degrees of bias, with 10 random experiments to increase statistical strength.

Table~\ref{tab:4} summarizes the TCAV scores for the doctor-nurse dataset with gender bias; we exchange columns from $D_M$ to $M_D$ to emphasize the change from subgroups to TCAV scores for men in the doctor class.
We included scores that are statistically insignificant from random because this does not imply independence between concept and target.
Metric 4 shows the proposed operational fairness metric, calculated by averaging across classes the difference in TCAV scores.
Although the scores follow expected trends by increasing for models that benefit from the concept bias and vice versa, neither they nor metric 4 estimated the degree of bias accurately.
This happens because images whose directional derivatives exceed a threshold does not necessarily receive the prediction for that reason, similar to how focusing on the face in Grad-CAM does not guarantee different predictions if gender is reversed.
This also insinuates that metric 4 already implicitly account for the baseline concept activation, by taking the difference between scores, yet still does not accurately reflect performance disparity.

\begin{table}[h]
\vskip 0.1in
\begin{center}
\begin{small}
\begin{sc}
\begin{tabular}{ c|c|cccc|c } 
 \toprule
 Ratio & Unfairness & $M_D$ & $F_D$ & $M_N$ & $F_N$ & M4 \\ 
 \midrule
 1:0 & 19.4 & 100 & 54 & 98 & 56 & 44.0 \\ 
 3:1 & 3.7 & 96 & 57 & 93 & 57 & 37.5 \\ 
 1:1 & 2.0 & 89 & 82 & 90 & 84 & 6.5 \\
 1:3 & 9.2 & 58 & 91 & 60 & 92 & 32.5 \\
 0:1 & 25.9 & 62 & 94 & 61 & 98 & 34.5 \\
 \bottomrule
\end{tabular}
\end{sc}
\end{small}
\end{center}
\vskip -0.1in
\caption{TCAV score (in percentage) between target classes and concepts and subgroup performance differences (unfairness) for the doctor-nurse dataset with gender bias. M4: class averages of differences in similarity.}
\label{tab:4}
\end{table}

\section{Discussion and Conclusion}
We summarize results and discuss if explanations help with bias detection by evaluating whether the aforementioned desirable characteristics were met. 

\subsection{Can One Detect Bias from Explanations}
There exists fundamental limitations on detecting bias through explanations.
If one does not associate certain features as indicating bias for Grad-CAM, or does not think of curating certain concepts for TCAV, then the quality of generated explanations is outweighted by the human element.
Furthermore, biases that affect all features in the image, such as image quality, lighting changes, and rotations, would go undetected since people expect to find an object; for TCAV, it is difficult to represent such biases using a set of images.

Although we observed explanations that suggest bias for biased models, the two mechanisms fall short of contributing meaningfully to bias detection in practice by poorly estimating the performance disparities caused by bias.
This implies that explanations could only detect very strong biases, when TCAV scores are statistically significant and many incorrect predictions focus on biased features for Grad-CAM.
Combined with the fact that unbiased models sometimes generate explanations that show bias, it is hard for researchers to decide whether the amount of bias present is acceptable for the application or how to mitigate the bias.

\subsection{Do Explanations Introduce Bias}
Introducing additional bias into explanations, not limited to the bias detection use case, decreases trustworthiness by incorrectly explaining why a model makes given decisions.
For Grad-CAM and saliency maps, this happens when humans misinterpret why a region is highlighted due to inherently biased assumptions. 
For TCAV, this comes from concepts replicating existing bias and then displaying unwanted correlations with target classes or negative examples.
In our experiments, the gender concepts contained more Caucasians, which resulted in higher scores for the volleyball but not basketball class by effectively measuring both race and gender together.
Detecting and mitigating bias during explanation generation require bias-aware experts making careful decisions. 

\subsection{Are Explanations Better Than Auditing}
Explanations perform worse than manual auditing and calculating unfairness in terms of estimating the degree of bias. 
Since examining Grad-CAM explanations require similar human effort as generating fine labels, the explanations only provide efficiency improvements when detecting multiple attributes simultaneously.
However, from our results, most of the explanations only focus on the most prominent bias, questioning the performance in this scenario.
We also discovered that TCAV concepts has limited re-usability, thus, the human effort spent is similar to curating the auditing dataset.
In conclusion, these results indicate that explanations do not have significant advantage over auditing in terms of efficiency, yet poorly estimates the degree of bias or even introduces additional bias into the analysis.
Our study suggests that devising explanation mechanisms for specific tasks, taking into account the intended use scenario, acceptable level of performance, and amount of human effort involved, would increase usefulness in practice. 

\subsection{Does Studying Bias Help Explanations}
Explanations provide meaningful summaries of the model's decision process by describing it in terms of human-understandable, usually higher-level, ideas.
Most bias belong to higher-level concepts as well.
Assuming that an unbiased model makes decisions based on a set of qualified features, then the biased model would likely incorporate additional features that represent the bias.
Comparing between explanations generated from two models with different degrees of bias ascertains whether explanations robustly reflect changes in the set of discriminating features.


\bibliography{example_paper}
\bibliographystyle{icml2020}

\end{document}